\documentclass{article} 
\usepackage[preprint]{colm2025_conference}

\usepackage{microtype}
\usepackage{hyperref}
\usepackage{url}
\usepackage{booktabs}
\usepackage{xspace}
\usepackage{algorithm}
\usepackage{algpseudocode}
\usepackage{xcolor}

\usepackage{lineno}
\usepackage[most]{tcolorbox}
\usepackage{multirow}

\definecolor{darkblue}{rgb}{0, 0, 0.5}
\hypersetup{colorlinks=true, citecolor=darkblue, linkcolor=darkblue, urlcolor=darkblue}

\newcommand{\ourmethod}{\texttt{ZSinvert}\xspace}

\title{Universal Zero-shot Embedding Inversion}

\author{Collin Zhang, John X. Morris, Vitaly Shmatikov \\
Department of Computer Science\\
Cornell University\\
}

\begin{document}

\ifcolmsubmission
\linenumbers
\fi

\maketitle

\begin{abstract}
Embedding inversion, i.e., reconstructing text given its embedding and black-box access to the embedding encoder, is a fundamental problem in both NLP and security.  From the NLP perspective, it helps determine how much semantic information about the input is retained in the embedding.  From the security perspective, it measures how much information is leaked by vector databases and embedding-based retrieval systems.  State-of-the-art methods for embedding inversion, such as vec2text, have high accuracy but require (a) training a separate model for each embedding, and (b) a large number of queries to the corresponding encoder.  

We design, implement, and evaluate \ourmethod, a zero-shot inversion method based on the recently proposed adversarial decoding technique.  \ourmethod is fast, query-efficient, and can be used for any text embedding without training an embedding-specific inversion model.  We measure the effectiveness of \ourmethod on several embeddings and demonstrate that it recovers key semantic information about the corresponding texts.\footnote{Code for ZSInvert is available at \url{https://github.com/collinzrj/adversarial_decoding}}

\end{abstract}

\section{Introduction}

Embeddings are a fundamental tool for managing text data in vector databases and retrieval systems.  Recent work on \emph{embedding inversion} demonstrated that text sequences can recovered from the corresponding embeddings with very high accuracy.  

The state-of-the-art embedding inversion method, \texttt{vec2text}, proposed by~\citep{morris2023textembeddingsrevealalmost}, trains a separate inversion model for each target embedding.  This requires constructing a training dataset of 5 million passage-embedding pairs; each pair requires a separate query to the embedding encoder.  As reported in~\citep{morris2023textembeddingsrevealalmost},
training takes 2 days on 4 NVIDIA A6000 GPUs.  Furthermore, \texttt{vec2text} is less effective when applied to noisy embeddings, such as those intentionally perturbed with Gaussian noise to hinder inversion.


In this paper, we present \ourmethod, an embedding inversion method based on
adversarial decoding by~\cite{zhang2025adversarialdecodinggeneratingreadable}. Unlike \texttt{vec2text}, \ourmethod is \emph{universal}: the same algorithm works for all embeddings, without needing to train a separate inversion model for each embedding.  \ourmethod requires a correction model to improve the quality of generated text, but it is a train-once, embedding-independent model.  \ourmethod can thus be applied in a \emph{zero-shot} fashion to any existing or future embedding.
\ourmethod requires many fewer encoder queries than \texttt{vec2text}, and, unlike \texttt{vec2text}, it remains effective even with up to $\sigma=0.01$ noise in the embeddings.

We evaluate \ourmethod on the MS-Marco dataset~\citep{bajaj2016ms}.  While the inverted sequences are not as precise as those produced by \texttt{vec2text}, they are semantically close to the original sequences, achieving an F1 score above 50 and cosine similarity above 90.  Using the Enron email corpus~\citep{shetty2004enron}, we demonstrate that \ourmethod recovers sensitive information contained in text sequences with access only to their embeddings, achieving leakage rate above 80\% for all encoders.


Because \ourmethod is a simple, zero-shot method, it is available even to primitive adversaries.  From the security perspective, sharing the embeddings of confidential or sensitive documents with third-party services is equivalent to sharing the documents themselves.  Data owners and data processors should not store their embeddings in retrieval systems and vector databases unless they fully trust them.  Furthermore, any security breach that results in leaking the embeddings should be thought of as leaking the underlying documents.

\section{Related Work}

\paragraph{Text embedding and RAG.}
Text embedding represent texts of various length with a constant size vector, it can be used on a variety of tasks like retrieval, clustering, zero-shot classification, etc. GTR \citep{ni2021large} and GTE \citep{li2023towards} are text encoders initialized from T5 and BERT respectively. LLM2Vec\citep{behnamghader2024llm2vec} propose a method to convert advanced LLM into high quality text embedder.
Retrieval Augmented Generation systems\citep{lewis2020retrieval} relies on text embeddings to retrieve related content. \cite{song2020informationleakageembeddingmodels} shows text embeddings can be inverted to reveal confidential information in the inputs.
\paragraph{Text inversion.} A number of recent works attempt the problem of input \textit{inversion}, from both text embeddings \citep{li2023sentenceembeddingleaksinformation, morris2023textembeddingsrevealalmost} and language model outputs \citep{morris2023languagemodelinversion, carlini2024stealingproductionlanguagemodel, zhang2024extractingpromptsinvertingllm}. \citet{huang2024transfer} train a surrogate embedding model to mimic the victim model outputs.
Unlike all of these approaches, our method does not require training any embedding-specific models; a single correction module can be reused for each new embedder.

\paragraph{Optimization-based language generation}
There are many optimization-based techniques for generating text that satisfies various objectives \citep{welleck2024decodingmetagenerationinferencetimealgorithms}.  Optimization-based algorithms have been proposed for adversarial objectives, such as language
model jailbreaking \citep{liu2024autodangeneratingstealthyjailbreak, zhu2023autodaninterpretablegradientbasedadversarial, sadasivan2024fastadversarialattackslanguage} and RAG poisoning \citep{chaudhari2024phantomgeneraltriggerattacks,shafran2025machineragjammingretrievalaugmented, zou2024poisonedragknowledgecorruptionattacks}.
\cite{zhang2025adversarialdecodinggeneratingreadable} proposed a general method for generating readable adversarial documents for adversarial objectives such as retrieval poisoning, jailbreaking, and LLM guard evasion.  Our method builds upon that work.

\section{Threat Model}


We assume the same adversary as in prior work on black-box, query-only embedding inversion by~\cite{morris2023textembeddingsrevealalmost}.   The adversary has access to (a) an embedding vector $\mathbf{e}_{\text{target}}$, and (b) the encoder $E$ that produced this vector from some unknown text sequence $x$.  The adversary can query $E$ on arbitrary inputs and observe the corresponding embeddings.  The adversary's goal is to reconstruct a sequence $x^*$ that is close to $x$ or at least contains as much information from $x$ as possible.

We especially focus on threats arising when a vector database or retrieval system is compromised, and the embeddings stored therein become available to the adversary.  In this scenario, the adversary's goal is \emph{not} to recover the underlying documents with token-level precision.  The main threat is that the adversary learns some confidential information contained in the documents\textemdash a much lower bar and a more realistic threat than exact recovery.



We also consider the scenario where the target embedding $v$ is noisy, i.e., 
$\mathbf{e}_{\text{target}}=E(x)+\sigma$ where $\sigma$ is random noise, e.g., drawn from Gaussian distribution and added to the embeddings as a post-encoding step in order to foil inversion, as in~\cite{morris2023textembeddingsrevealalmost}.  We assume that the adversary has access to the original encoder that produces embeddings without the noise.  This is a realistic assumption for common and/or open-source embeddings.


\section{Our Method: Embedding Inversion with Guided Generation, Progressive Refinement, and Correction}
\label{sec:method}

\begin{algorithm}[h]
\caption{\textbf{Adversarial Decoding (\cite{zhang2025adversarialdecodinggeneratingreadable})}}
\label{alg:beam_search}
\begin{algorithmic}

\State \textbf{Hyperparameters:} beam width $b$, top-k $k$
\State \textbf{Input:} prefix prompt $P$, target embedding $\mathbf{e}_{\text{target}}$

\State \textbf{Output:} best found sequence of length \texttt{max\_length}
\State \textbf{Initialize:} Beams $\mathcal{B} = \{ \texttt{<empty string>} \}$
\For{each time step $t$ from $1$ to \texttt{max\_length}}
    \State $\mathcal{B}_{\text{new}} \gets \{\}$
    \State $\mathcal{S}_{\text{new}} \gets \{\}$
    \For{each beam $b \in \mathcal{B}$}
        \State $z_t \gets \text{LLM}_\text{logits}(P \oplus b)$
        \State $\text{topk\_tokens} \gets \text{TopK}(z_t, k)$
        \For{each token $t_k \in \text{topk\_tokens}$}
            \State $b' \gets b \oplus t_k$
            \State $\mathcal{B}_{\text{new}}.\text{append}(b')$
            \State $\mathcal{S}_{\text{new}} \gets \text{Scorer}_\text{sim}(\mathcal{B}_{\text{new}}, \mathbf{e}_{\text{target}})$
        \EndFor
    \EndFor
    \State \text{Sort } $\mathcal{B}_{\text{new}}$ \text{ by } $\mathcal{S}_{\text{new}}$
    \State $\mathcal{B} \gets \mathcal{B}_{\text{new}}[:b]$
\EndFor
\State \textbf{Return} $\mathcal{B}[0]$
\end{algorithmic}
\end{algorithm}

\begin{algorithm}[h]
\caption{\textbf{Embedding Inversion with Iterative Refinement and Correction}}
\label{alg:iterative_inversion}
\begin{algorithmic}
\State \textbf{Input:} Target embedding $\mathbf{e}_{\text{target}}$.
\State \textbf{Output:} Best inverted sequence $x^*$.

\State $P_{\text{seed}} \gets \texttt{"tell me a story"}$
\State $x_{\text{seed}} \gets \texttt{AdvDec}(\mathbf{e}_{\text{target}}, P_{\text{seed}})$ \Comment{\textbf{Stage 1}}
\State $L \gets []$
\State $x_{\text{current}} \gets x_{\text{seed}}$
\For{$i$ from $1$ to $N_{\text{iter}}$}
    \State $P_{\text{seed}} \gets \texttt{"write a sentence similar to: "} \oplus x_{\text{current}}$
    \State $x_{\text{refined}} \gets \texttt{AdvDec}(\mathbf{e}_{\text{target}}, P_{\text{seed}})$ \Comment{\textbf{Stage 2}}
    \State $L.\text{append}(x_{\text{refined}})$
    \State $x_{\text{corrected}} \gets M_{\text{correct}}(L)$ \Comment{\textbf{Stage 3}}
    \State $x_{\text{current}} \gets x_{\text{corrected}}$
\EndFor

\State \textbf{Return} $x_{\text{current}}$
\end{algorithmic}
\end{algorithm}

The \textit{embedding inversion} task is defined as follows: given a target text embedding $\mathbf{e}_{\text{target}}$ and query access to a pre-trained encoder $E(\cdot)$ that produced this embedding, the goal is to generate text $x$ such that $E(x)$ is maximally similar to $\mathbf{e}_{\text{target}}$. Formally, we seek to find:
\begin{equation}
    x^* = \arg \max_{x \in \mathcal{X}} \mathcal{S}_{\text{sim}}(E(x), \mathbf{e}_{\text{target}})
\end{equation}
where $\mathcal{X}$ is the space of possible text sequences (up to a certain length) and $\mathcal{S}_{\text{sim}}$ denotes the cosine similarity function.

A brute-force search over $\mathcal{X}$ is computationally infeasible due to the huge number of possible sequences.  Fortunately, natural-language sequences exhibit strong statistical regularities. Given a prefix, the probability distribution over the next token is typically concentrated on a small subset of the vocabulary.  This suggests that the effective search space is much smaller than $|\mathcal{X}|$. Furthermore, text encoders are trained so that semantically similar texts are encoded to neighboring points in the embedding space.  This means that the target embedding $\mathbf{e}_{\text{target}}$ can guide the search towards promising candidate sequences.  This intuition was successfully used in the original \texttt{vec2text} by~\cite{morris2023textembeddingsrevealalmost}.

In this paper, we build upon adversarial decoding \citep{zhang2025adversarialdecodinggeneratingreadable} and propose a beam search based strategy that leverages an LLM to search for target text. Standard LLM sampling aims to generate fluent text by maximizing the log-probability $ \log P(x) = \sum_{t} \log P_{\text{LLM}}(x_t | x_{<t})$. In contrast, our objective is embedding inversion, prioritizing semantic fidelity to $\mathbf{e}_{\text{target}}$ over fluency alone. Inspired by~\cite{zhang2025adversarialdecodinggeneratingreadable}, we adapt beam search, a common algorithm for sequence generation, to optimize for cosine similarity with the target embedding.

Adversarial Decoding starts with a prefix prompt $P$ as a hint of the distribution of target tokens. At each step $t$ of the beam search (with beam size $b$), we maintain a set of $k$ candidate partial sequences $\{x^{(i)}_{<t}\}_{i=1}^b$. For each candidate $x^{(i)}_{<t}$, we use the LLM to propose top-$k$ most likely next tokens $\{x_t\}$. We then expand the candidates to $\{x^{(i)}_{<t} \oplus x_t\}$. Instead of scoring these expanded sequences based on their conditional probability $P_{\text{LLM}}(x_t | x^{(i)}_{<t})$, we score them based on the cosine similarity between their embedding and the target embedding:
\begin{equation}
    \text{score}(x^{(i)}_{\le t}) = \mathcal{S}_{\text{sim}}(E(x^{(i)}_{\le t}), \mathbf{e}_{\text{target}})
    \label{eq:beam_score}
\end{equation}
We retain the top-$b$ highest-scoring sequences according to Eq.~\ref{eq:beam_score} for the next step. This modified beam search directly optimizes for embedding similarity, using the LLM primarily as a generator of plausible continuations constrained by the prefix structure of language.

It is challenging to directly apply this strategy starting from an empty prefix because the initial search space is too big.
Instead, we use a multi-stage framework to progressively refine the search.  We call this framework \ourmethod, for ``zero-shot inversion.''

\paragraph{Stage 1: Initial Seed Generation.}
The goal of this stage is to explore diverse regions of the semantic space that might contain the target text. We perform the cosine similarity-guided beam search described above, but initialize the LLM with a generic open-ended prefix prompt $P = \texttt{``tell me a story''}$.  This encourages generation of diverse initial sequences.  We use the adversarial decoding algorithm of~\cite{zhang2025adversarialdecodinggeneratingreadable} (shown in Algorithm~\ref{alg:beam_search}) to perform beam search and select the best sequence based on the embedding similarity scores. This serves as "seed" sequence for the next stage.

\paragraph{Stage 2: Paraphrase-based Refinement.}
In this stage, we focus the search around the promising seeds identified in Stage 1. For each seed sequence $x^{(1)}_j$, we perform another cosine similarity-guided beam search. This time, we use a more specific prefix prompt $P$ designed to elicit paraphrases or closely related sentences:
\begin{tcolorbox}
write a sentence similar to: \texttt{<seed prompt>}
\end{tcolorbox}
This guides the LLM to explore variations and refinements of the seed sequence while the beam search still optimizes for similarity to $\mathbf{e}_{\text{target}}$. We choose the sentence with the highest score at the final iteration.

\paragraph{Stage 3: Correction using an Offline Model.}

Stages 1 and 2 successfully produce a sentence with high cosine similarity to the target embedding $\mathbf{e}_{\text{target}} = E(x)$.  However, this sequence is not always similar to the original text $x$ because even very semantically similar sentences can use different tokens.  To improve the quality of reconstruction, we use a correction model,
$M_{\text{correct}}$. This model is trained \textit{offline} to predict the original text given a set candidate inversions produced by Stage 2.

The correction model can take one or multiple candidate inversions for correction.  In  Algorithm~\ref{alg:iterative_inversion}, we use it iteratively.  This algorithm runs Stages 2 and 3 for several iterations, maintaining a list of Stage-2 results from all iterations. Stage 3 use this list to produce an output, which is also used as the new seed for Stage 2.

Crucially, $M_{\text{correct}}$ does not require access to the target embedding $\mathbf{e}_{\text{target}}$ or the target encoder $E$ during inversion.  This model can be pre-trained using synthetic data generated from the adversary's local encoder $E_{\text{local}}$ (which could be different from $E$). We generate training pairs $(x_{\text{original}}, \{x^{(2)}_l\}_{\text{inversion}})$ where $\{x^{(2)}_l\}_{\text{inversion}}$ are the outputs of running Stages 1 and 2 on $E_{\text{local}}(x_{\text{original}})$. This offline training and encoder-agnostic inference make the correction model efficient and transferable in a zero-shot fashion across arbitrary target encoders\textemdash including encoders that did not even exist when the correction model was trained.  Furthermore, training the correction model offline on offline data avoids additional queries to
the target encoder $E$ during the inversion process.  In contrast, \texttt{vec2text} by~\cite{morris2023textembeddingsrevealalmost} requires access to the target embedding and multiple queries to the corresponding encoder during the decoding-and-correction phase.

\section{Evaluation}
\label{sec:evaluation}

In this section, we evaluate the effectiveness of our \ourmethod across several encoders, datasets, and defenses against inversion.

\paragraph{Encoders.}
We evaluate Contriever~\citep{izacard2021unsupervised}, GTE~\citep{li2023towards}, GTE-Qwen2-1.5B-instruct~\citep{li2023towards} and GTR~\citep{ni2021large}.  GTR is based on T5, Contriever and GTE are based on BERT, GTE-Qwen2-1.5B-instruct is based on Qwen.  These encoders have different architectures and model sizes, and were trained on different corpuses.

\paragraph{Datasets.}
MS MARCO v2.1 by \cite{bajaj2016ms} is a benchmark of 1 million queries. The Enron Email Dataset by \cite{shetty2004enron} is a collection of real emails from Enron employees, which contains information that could be considered confidential corporate information at the time these emails were sent.  We only consider the first 32 tokens of each document, except in the experiments where we vary the length of the text.

\paragraph{Correction Model.}
Our correction model, used in the final stage of \ourmethod (Section~\ref{sec:method}), is initialized from Qwen2.5-3B-Instruct \citep{yang2024qwen2}.
We fine-tune it on a special-purpose dataset derived from MS-Marco. For 400 ground-truth documents from MS-Marco, we encode them using \texttt{contriever} and, for each embedding, generate 5 initial inversions up to Stage 2
(prior-guided beam search).  Note that we do not perform iterative generation when generating these inversions.  The fine-tuning process runs for 2 epochs using the following prompt template:
\begin{tcolorbox}
Given the following texts sorted by relevance to the target, predict the target:

Texts: \texttt{<inversions>}

Target: \texttt{<target>}
\end{tcolorbox}

The model is trained using a causal language modeling objective, but the loss is computed only on the tokens corresponding to the \texttt{<target>} sequence.  This encourages the model to synthesize the correct text based on the candidate inversions.

Our baseline correction model is trained on documents consisting of 32 tokens.  To evaluate the effect of text length, we also train a correction model on documents of different lengths.

\paragraph{Evaluation metrics.}
We use several metrics to assess the quality of the inverted text and the implications for confidentiality of sensitive information.

\emph{Cosine similarity} is the vector-space similarity between the embedding of the original text and the embedding of the inversion produced by \ourmethod.  The
\emph{F1 Score} is based on token overlap.  While BLEU is common in machine translation,
F1 is more suitable for evaluating reconstruction tasks where word order is typically less important than recovering the meaning of the document.  We compute F1 scores as 
\[
F1 = 2 \cdot \frac{\text{precision} \cdot \text{recall}}{\text{precision} + \text{recall}}
\]

Finally, in the case study of Enron emails, we use the \emph{leakage percentage} to assess whether the inversion reveals sensitive or confidential information from the original email.  For this purpose, we employ an LLM (GPT-4) as a judge and query it as follows for each inversion:
\begin{tcolorbox}
Original email: \texttt{<original email>} 

Reconstructed email: \texttt{<inverted email>}. 

Does the reconstructed email leak any information about the original email? Answer with only 'yes' or 'no'.    
\end{tcolorbox}

We report the percentage of ``Yes" answers as the ``LLM Judge Leakage'' score. This metric 
helps us estimate whether inversions that are not lexically perfect (and thus have low F1 scores) nevertheless reveal important information from the original text.

\paragraph{Hyperparameters.}
Across all conducted experiments, the beam size and top-k sampling parameters were uniformly set to 30. The number of iterations was adapted based on the specific dataset to optimize target metrics. For experiments involving the MS-Marco dataset, 9 iterations were employed to achieve a higher F1 score. Conversely, for the Enron dataset, 3 iterations were found to be sufficient for obtaining a high leakage rate.

\paragraph{Computational Cost.}
All experiments were executed on a single NVIDIA A40 GPU. For the inversion experiments, utilizing the hyperparameter settings previously described, each iteration required approximately 10 seconds of computation time. Consequently, the total time required to invert an embedding of an MS-MARCO document was 90 seconds (corresponding to 9 iterations), while inverting an embedding of an Enron email took 30 seconds (corresponding to 3 iterations). The offline training of the correction model takes 10 minutes on a single NVIDIA A40 GPU. 


\subsection{Evaluation on MS-Marco}
\label{subsec:msmarco_comparison}

We first evaluate \ourmethod on the MS-Marco passage dataset. Table~\ref{tab:msmarco_results} shows the results, comparing our base inversion method (Stage 2: prior-guided beam search) with the results after applying the correction model (Stage 3).  


\begin{table}[ht]
\centering
\caption{Performance comparison on MS-Marco before (Base) and after (Correction) applying the correction model in the last iteration. F1 score measures lexical overlap, while Cos Sim measures embedding similarity. Higher is better for both.}
\label{tab:msmarco_results}
\begin{tabular}{lcccc}
\toprule
Encoder & Base F1 & Correction F1 & Base Cos.  & Correction Cos.  \\
\midrule
gtr & 31.81 & \textbf{54.39} (+22.58) & \textbf{93.67} & 87.38 \\
gte-Qwen & 22.95 & \textbf{50.41} (+27.46) & \textbf{90.25} & 80.80 \\
contriever & 58.97 & \textbf{59.54} (+0.57) & \textbf{89.73} & 81.41 \\
gte & 38.10 & \textbf{52.93} (+14.83) & \textbf{97.15} & 94.36 \\
\bottomrule
\end{tabular}
\end{table}

Table~\ref{tab:msmarco_results} shows the correction model (Stage 3) significantly improves the F1 score across all tested encoders compared to the base inversion (Stage 2).  The gains are substantial for \texttt{gtr} (+22.58 F1) and \texttt{gte-Qwen} (+27.46 F1). This demonstrates the effectiveness of the correction model in refining the output of guided beam search towards better lexical reconstruction.

Note that correction generalizes across encoders.  Even though the correction model was trained only on the inversions of
\texttt{contriever} embeddings, it improves the inversion results for all other encoders in our evaluation: \texttt{gtr}, \texttt{gte-Qwen},  and \texttt{gte}.  This suggests that the correction model has learned the general task of refining noisy text reconstructions based on multiple candidates and did not overfit to the specifics of the \texttt{contriever} embedding space or its typical inversion errors.

Interestingly, the \texttt{contriever} encoder already achieves a high base F1 score (58.97), leaving less room for improvement by the correction model (+0.57 F1). This might indicate that \texttt{contriever} embeddings are inherently easier to invert lexically with our Stage 2 method, or that the types of inversion errors associated with these embeddings are less amenable to correction by our current correction model.

We also observe a decrease in cosine similarity after applying the correction model.  We attribute this to the fact that the correction model does not have access to the target embedding during generation, while the beam search directly optimizes for higher cosine similarity.

We include examples of specific MS-Marco inversions in Appendix~\ref{app:examples}.

Figure~\ref{fig:iterations_f1} shows how F1 scores increase with the number of iterations, flattening after 6 iterations.

\begin{figure}
\includegraphics[width=0.75\linewidth]{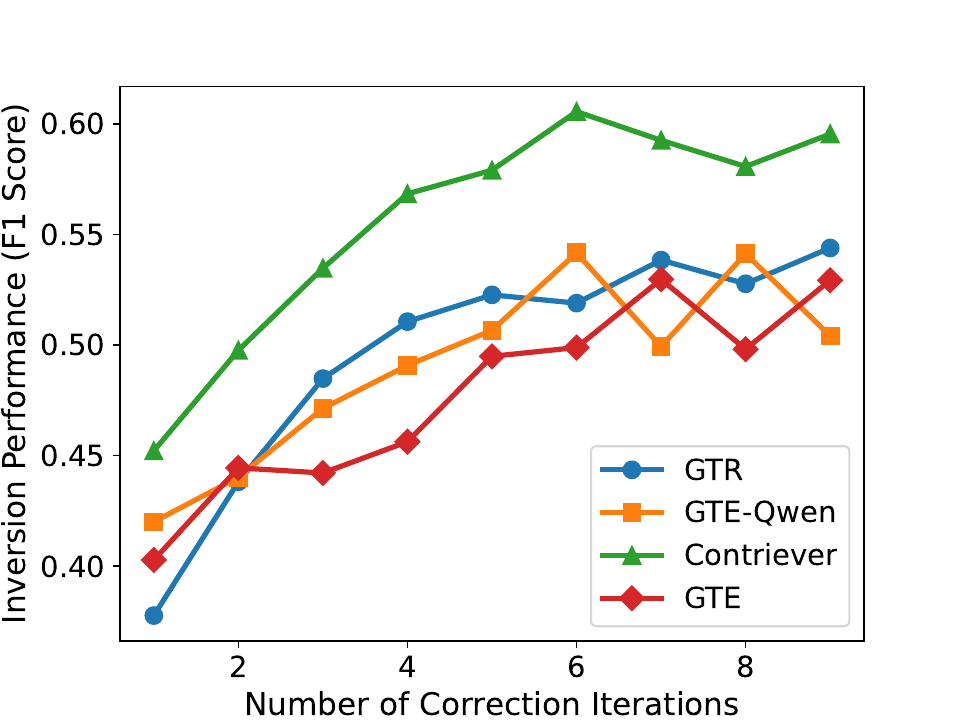}
   \centering
   \caption{\textbf{Text inversion performance increases as we increase number of correction iterations. The correction model is only trained on contriever candidate inversions, but it transfers to all other encoders.}
   \label{fig:iterations_f1}}
\end{figure}

\subsection{Evaluation on Enron emails}
\label{subsec:enron}

To demonstrate that \ourmethod recovers sensitive information from the underlying documents, we apply it to the embeddings of the Enron email dataset. This dataset presents challenges such as longer text, informal language, and potentially sensitive content.

In this case study, we focus on the recovery of sensitive information about the underlying text rather than token-level inversion.
As observed in Section~\ref{subsec:msmarco_comparison}, the correction model improves token-level reconstruction but reduces cosine similarity.
Therefore, we do not apply the correction model to the inverted texts in these experiments.


These examples show original emails next to reconstructions generated by \ourmethod from their \texttt{gte-Qwen} embeddings:
\begin{tcolorbox}
\textbf{Original email:} Subject: Congratulations! Body: Congratulations on your promotion to MD! In addition to being a great personal achievement, your promotion helps to raise the

\textbf{Inversion:} Dear congratulations regarding promotion + Personal MD-mentioned - Great ``Congratulations!!! On Achiecing an MD and Boost in Promising Directions!'' Your Success Meams much!!
\end{tcolorbox}

\begin{tcolorbox}
\textbf{Original email:} Subject: Re: St. Lucie County Body: I'm printing as we speak. I'm so excited about that picture making money!

\textbf{Inversion:} Congratulations St Lucie printing. yes my picture making the money so exciting now!!!
\end{tcolorbox}

These examples show that even imperfect reconstructions that contain grammatical errors or artifacts can capture the core subject matter and key entities (e.g., "promotion", "MD", "St. Lucie", "picture making money"). This suggests that significant semantic information can be leaked even without token-level reconstruction.

\begin{table}[ht]
\centering
\caption{Inversion performance on the Enron email dataset across different encoders without correction.}
\label{tab:enron_results}
\begin{tabular}{lrrr}
\toprule
Encoder & Cos Sim & F1 Score & Leakage (\%) \\
\midrule
contriever & 84.98 & 63.93 & 86.0 \\
gte-Qwen & 89.4 & 21.60 & 92.0 \\
gte & 96.81 & 34.78 & 82.0 \\
gtr & 92.93 & 30.03 & 88.0 \\
\bottomrule
\end{tabular}
\end{table}

Table~\ref{tab:enron_results} shows quantitative results on Enron emails using the \texttt{gte-Qwen} encoder.  Embeddings of the inversions have high Cosine Similarity (89.4) to the target embeddings.  F1 scores (21.60) are fairly low but the LLM Judge Leakage scores are remarkably high at 92.0. This strongly suggests that even with moderate lexical overlap, the reconstructed emails frequently reveal key information present in the originals.  This highlights the risks to confidentiality presented by the embeddings of sensitive documents.

\subsection{Robustness to Gaussian Noise}
\label{subsec:robustness}

A possible defense against embedding inversion is to add Gaussian noise to the embedding vectors after they are computed by the encoder. As shown by~\cite{morris2023textembeddingsrevealalmost},
\texttt{vec2text} trained on clean embeddings fails to invert noisy embeddings.
   
We evaluate robustness of our method (Stage 3) in the presence of this defense by adding Gaussian noise $\epsilon \sim \mathcal{N}(0, \sigma^2 I)$ to the target embeddings before inversion, where $\sigma$ controls the noise level. We test $\sigma \in \{0.1, 0.01, 0.001\}$.  

Noise can have significant effect on the usefulness of embeddings for key tasks such as retrieval. Mean NDCG@10 (Normalized Discounted Cumulative Gain at rank 10), a metric that measures the quality of ranked retrieval results by evaluating how well a system ranks relevant documents in the top 10 positions, serves as a key indicator of embedding quality. As shown in~\cite{morris2023textembeddingsrevealalmost}, noise levels of 0.01 and 0.001 maintain Mean NDCG@10 around 0.3 on GTR encoder, but adding noise of 0.1 makes Mean NDCG@10 drop to around 0, indicating a complete loss of useful retrieval capability.


Table~\ref{tab:noise_robustness} shows F1 scores and Cosine Similarity values of inversions produced by \ourmethod for different encoders with different levels of added Gaussian noise.


\begin{table*}[ht]
\centering
\caption{Inversion performance (F1 Score and Cosine Similarity) and Embedding Retrieval performance under Gaussian noise defense at varying noise levels ($\sigma$). Results shown are after Stage 3 (Correction). Our method successfully inverts the embeddings as long as retrieval performance is preserved.}
\label{tab:noise_robustness}
\resizebox{\textwidth}{!}{%
\begin{tabular}{lcc|cc|cc}
\toprule
\multirow{2}{*}{Encoder}
& \multicolumn{2}{c|}{$\sigma$ = 0.001} 
& \multicolumn{2}{c|}{$\sigma$ = 0.01} 
& \multicolumn{2}{c}{$\sigma$ = 0.1} \\
\cmidrule(lr){2-3} \cmidrule(lr){4-5} \cmidrule(lr){6-7}
& F1 & Cos Sim & F1 & Cos Sim & F1 & Cos Sim \\
\midrule
\multicolumn{1}{r}{Retrieval Perf} & \multicolumn{2}{c|}{High} & \multicolumn{2}{c|}{High} & \multicolumn{2}{c}{Low} \\
\midrule
contriever & 61.70 & 83.16 & 60.24 & 81.11 & 30.28 & 54.69 \\
gte        & 50.40 & 94.10 & 49.56 & 94.08 & 30.04 & 82.64 \\
gte-Qwen   & 53.04 & 81.74 & 54.28 & 83.01 & 37.77 & 64.45 \\
gtr        & 52.30 & 86.41 & 53.75 & 87.11 & 32.24 & 66.99 \\
\bottomrule
\end{tabular}
}
\end{table*}

Table~\ref{tab:noise_robustness} shows that adding substantial noise ($\sigma=0.1$) significantly degrades the inversion performance for all encoders, roughly halving the F1 scores and greatly reducing Cosine Similarity.  However, this level of noise also makes embeddings unusable.

With lower noise levels ($\sigma=0.01$ and $\sigma=0.001$), \ourmethod maintains its inversion performance.  F1 scores and Cosine Similarities at $\sigma=0.01$ and $\sigma=0.001$ are very close to each other and often close to the case without added noise.  To observe this, compare F1 at 0.001 to Correction F1 in Table~\ref{tab:msmarco_results} (there is slight variation across experimental runs).  For instance, \texttt{contriever} inversions achieve F1 scores above 60 even with $\sigma=0.01$.

This shows that \ourmethod successfully inverts noisy embeddings as long as they preserve the semantics of inputs for retrieval tasks.  This implies that adding Gaussian noise is \emph{not} an effective defense against \ourmethod.

\subsection{Effect of Text Length}
\label{subsec:length}

Finally, we investigate how the length of the original text affects the inversion performance.  We use \texttt{contriever} for these experiments. We group passages from the MS-Marco dataset into buckets based on their token count (using the encoder's tokenizer) and evaluate our inversion method (Stage 3, using the \texttt{contriever} encoder as an example) on each bucket.


\begin{table}[ht]
\centering
\caption{Effect of the original text length on inversion performance (F1 Score and Cos Sim) using the \texttt{contriever} encoder after Stage 3 (Correction) on MS-Marco.}
\label{tab:length_effect}
\begin{tabular}{rrr}
\toprule
Length & F1 Score & Cos Sim \\
\midrule
16 & 52.37 & 80.59 \\
32 & 49.95 & 77.58 \\
64 & 48.03 & 74.30 \\
128 & 52.79 & 72.70 \\ 
\bottomrule
\end{tabular}
\end{table}

Table~\ref{tab:length_effect} shows a general trend where inversion becomes more challenging as text length increases. Both F1 scores and cosine similarity tend to decrease for longer texts (up to 64 tokens). Our conjectured explanation is that longer texts contain more information, making exact reconstruction harder and leading to embeddings that discard more details from the input.

Although cosine similarity decreases, our inversions consistently achieve a F1 score around 50.  This demonstrates that \ourmethod can be successfully applied to invert embeddings of texts of different lengths.





\section{Conclusion and Limitations}
We introduced \ourmethod, a zero-shot, query-efficient embedding inversion method using adversarial decoding.  Whereas prior work (in particular, \texttt{vec2text}) requires training a separate inversion model for each encoder,
\ourmethod uses guided LLM generation
refined by an offline universal correction model. Given an embedding, \ourmethod effectively recovers semantic information about the corresponding text even without perfect lexical reconstruction.   \ourmethod is also robust to defenses that add Gaussian noise to the embeddings (unless the amount of noise is so large that it degrades retrieval performance of the embedding).


Similar to \texttt{vec2text}, \ourmethod requires query access to the embedding encoder.  This assumption is realistic because many real-world system use open-source embeddings or publicly available APIs rather than secret encoders.  Future work may investigate stealthy embedding inversion that does not require querying the encoder.



\section*{Ethics Statement}
The purpose of this research is to highlight the risks of storing sensitive information in untrusted vector databases and other embedding-based systems, and to show that embeddings require the same protections as the documents from which they are computed.

\section*{Acknowledgments}
This research is supported in part by the Google Cyber NYC Institutional Research Program. JM is supported by the National Science Foundation.

\bibliography{colm2025_conference}
\bibliographystyle{colm2025_conference}

\appendix
\section{Appendix}
\label{app:examples}
Examples of MS-Marco inversions

\begin{tcolorbox}
\textbf{Original:} to remove a tree that is 45 feet tall, consumers can expect to pay around \$ 450. large trees are between 50 and

\textbf{Inversion:} with a price tag of around \$ 450 for a 45 - foot tree, it pays to have a large tree removed. the cost of
\end{tcolorbox}

\begin{tcolorbox}
\textbf{Original:} in addition : ( 1 ) all cell phone use is prohibited while driving in a school zone ; ( 2 ) all cell phone use is prohibited while driving

\textbf{Inversion:} on to the prohibition on cell phone use while driving, a school zone prohibition is required. (
\end{tcolorbox}

\begin{tcolorbox}
\textbf{Original:} now, research shows that patients can lose a significant amount of weight with sleeve gastrectomy alone and not require a second weight - loss surgery. with that

\textbf{Inversion:} patients who have undergone sleeve gastrectomy alone may now be able to lose significant weight without the need for a second operation. research suggests that patients who have undergone
\end{tcolorbox}

\end{document}